# Non-iterative Coarse-to-fine Registration based on Single-pass Deep Cumulative Learning

Mingyuan Meng[1], Lei Bi[1], Dagan Feng[1,2], and Jinman Kim[1]

[1] School of Computer Science, The University of Sydney, Sydney, Australia.
[2] Med-X Research Institute, Shanghai Jiao Tong University, Shanghai, China.
`lei.bi@sydney.edu.au`

**Abstract.** Deformable image registration is a crucial step in medical image analysis for finding a non-linear spatial transformation between a pair of fixed and moving images. Deep registration methods based on Convolutional Neural Networks (CNNs) have been widely used as they can perform image registration in a fast and end-to-end manner. However, these methods usually have limited performance for image pairs with large deformations. Recently, iterative deep registration methods have been used to alleviate this limitation, where the transformations are iteratively learned in a coarse-to-fine manner. However, iterative methods inevitably prolong the registration runtime, and tend to learn separate image features for each iteration, which hinders the features from being leveraged to facilitate the registration at later iterations. In this study, we propose a Non-Iterative Coarse-to-finE registration Network (NICE-Net) for deformable image registration. In the NICE-Net, we propose: (i) a Single-pass Deep Cumulative Learning (SDCL) decoder that can cumulatively learn coarse-to-fine transformations within a single pass (iteration) of the network, and (ii) a Selectively-propagated Feature Learning (SFL) encoder that can learn common image features for the whole coarse-to-fine registration process and selectively propagate the features as needed. Extensive experiments on six public datasets of 3D brain Magnetic Resonance Imaging (MRI) show that our proposed NICE-Net can outperform state-of-the-art iterative deep registration methods while only requiring similar runtime to non-iterative methods.

## 1 Introduction

Deformable image registration is a fundamental requirement for a variety of clinical tasks such as tumor growth monitoring and organ atlas creation [1]. It aims to find a non-linear spatial transformation between a pair of fixed and moving images, which warps the moving image to align with the fixed image. Traditional registration methods address deformable registration as an iterative optimization problem. However, iterative optimization is time-consuming, especially for high-resolution 3D images such as brain Magnetic Resonance Imaging (MRI) [2], which limits its clinical applications as fast registration is widely desired in clinical practice. Recently, deep registration methods based on Convolutional Neural Networks (CNNs) have been widely used to perform fast and end-to-end registration in a non-iterative manner [3]. However, non-iterative deep registration methods usually work well for image pairs with small deformations while having degraded performance for large deformations [4, 5].

Iterative deep registration methods have been proposed to alleviate this limitation and are regarded as the state-of-the-art [6-10], in which the registration is performed by iteratively warping the moving image in a coarse-to-fine manner. Iterative coarse-to-fine registration usually is implemented by using multiple cascaded networks [6-9]. The first network performs coarse registration at the beginning and each following network is used to refine the registration based on the warped image derived from its former network. For example, Zhao et al. [8] proposed a Recursive Cascaded Network (RCN), where multiple CNNs were cascaded and were trained end-to-end. Mok et al. [9] proposed a Laplacian pyramid Image Registration Network (LapIRN), in which multiple CNNs at different pyramid levels were cascaded. However, these methods rely on multiple networks to perform coarse-to-fine registration, which inevitably raises a huge requirement for GPU memory. Recently, Shu et al. [10] proposed to use a single network (ULAE-net) to perform iterative coarse-to-fine registration. At each iteration, the ULAE-net produces a transformation to warp the moving image, and then the warped image is fed into the ULAE-net again to perform finer registration at the next iteration. However, these iterative deep registration methods all have certain limitations: (i) iterative learning inevitably increases computational loads and prolongs the registration runtime, and (ii) iterative methods usually learn separate image features for each iteration, which hinders the features from being leveraged at later iterations and also adds extra computational loads due to repeated feature learning. Hu et al. [11] proposed a Dual-stream Pyramid Registration Network (Dual-PRNet), in which coarse-to-fine registration is performed in a non-iterative manner. The Dual-PRNet, through warping image feature maps, can produce coarse-to-fine transformations within one iteration. However, its registration accuracy is unable to match iterative deep registration methods. In addition, few coarse-to-fine registration methods impose constraints on the transformations to keep their invertibility. Mok et al. [9] attempted to impose diffeomorphic constraints on the transformations in the LapIRN. However, this dramatically degraded the registration accuracy.

In this study, we propose a Non-Iterative Coarse-to-finE registration Network (NICE-Net) for deformable image registration. Compared to the state-of-the-art iterative deep registration methods, our NICE-Net can perform more accurate registration with a single network in a single iteration. The technical contributions of our NICE-Net are in two folds: (i) we propose a Single-pass Deep Cumulative Learning (SDCL) decoder that can cumulatively learn coarse-to-fine transformations within a single (iteration) pass of the network, and (ii) we propose a Selectively-propagated Feature Learning (SFL) encoder that can learn common image features for the whole coarse-to-fine registration process and selectively propagate the features as needed. We also incorporated penalizing negative Jacobian determinants into the loss function in coarse-to-fine registration, which allows to keep the transformation invertibility with a marginal degradation on the registration accuracy. We performed comprehensive experiments on six public datasets of 3D brain MRI.

## 2    Method

Image registration aims to find a spatial transformation $\phi$ that warps a moving image $I_m$ to a fixed image $I_f$, so that the warped image $I_m \circ \phi$ is spatially aligned with the fixed image $I_f$. In this study, the moving image $I_m$ and fixed image $I_f$ are two volumes defined in a 3D spatial domain $\Omega \subset \mathbb{R}^3$, and the $\phi$ is parameterized as a displacement field, following [4]. For coarse-to-fine registration settings, we define $L$ as the number of registration steps. At the $i^{th}$ step ($i \in \{1, 2, ..., L\}$), a transformation $\phi_i$ is produced with the $\phi_1$ as the coarsest transformation and the $\phi_L$ as the finest transformation. It should be noted that, for existing iterative methods [8-10], the $L$ also is the number of the cascaded networks (or running iterations), as they implement each registration step with a separate network (or iteration). However, our NICE-Net can implement all $L$ registration steps within one iteration. In addition, we create two image pyramids by downsampling the $I_f$ and $I_m$ with trilinear interpolation to obtain $I_f{}^i \in \{I_f{}^1, I_f{}^2, ..., I_f{}^L\}$ and $I_m{}^i \in \{I_m{}^1, I_m{}^2, ..., I_m{}^L\}$, where the $I_f{}^i$ and $I_m{}^i$ are the downsampled $I_f$ and $I_m$ by a factor of $0.5^{(L-i)}$ with $I_f{}^L = I_f$ and $I_m{}^L = I_m$.

The architecture of the proposed NICE-Net is illustrated in Fig. 1, which consists of a SFL encoder (detailed in Section 2.1) and a SDCL decoder (detailed in Section 2.2). The SFL encoder extracts image features and selectively propagates the features to the SDCL decoder through skip connections. The SDCL decoder performs $L$-step coarse-to-fine registration.

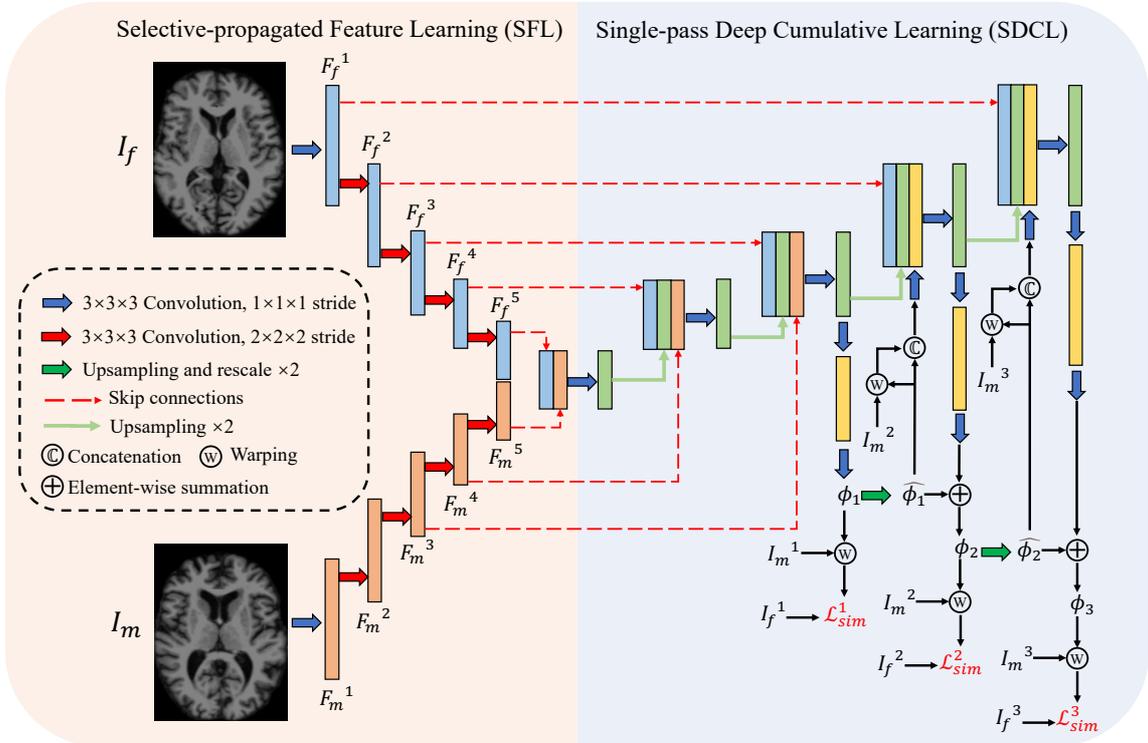

**Fig. 1.** Architecture of the NICE-Net. The registration step, $L$, is set as 3 for illustration.

## 2.1 Selectively-propagated Feature Learning (SFL)

The SFL encoder is a dual-path encoder with selective feature propagation. Unlike the existing deep registration methods that usually learn coupled image features from the concatenated fixed/moving images [4, 5, 8, 9], the SFL encoder learns features from the $I_f$ and $I_m$ separately using two paths. This dual-path design can derive uncoupled features of $I_f$ and $I_m$, which enables us to selectively propagate different features to perform different steps of coarse-to-fine registration. For example, the features of $I_f$ should always be propagated, as the $I_f$ is the registration target for all registration steps. However, the features of $I_m$ are important for the first step but not necessary for later, because the later steps refine the transformation based on the intermediary warped image (derived from its former step) rather than based on the original $I_m$. This selective feature propagation allows the network to learn common image features for the whole coarse-to-fine registration process without the need for repeated feature learning for each registration step.

Specifically, the SFL encoder has two identical, weight-shared paths, $P_f$ and $P_m$, which takes $I_f$ and $I_m$ as input, respectively. Each path consists of five successive $3 \times 3 \times 3$ convolutional layers, followed by LeakyReLU activation with parameter 0.2. Except for the first convolutional layer, each convolutional layer has a stride of 2 to reduce the resolution of feature maps. Through the SFL encoder, we can obtain two 5-level feature pyramids, $F_f{}^i \in \{F_f{}^1, F_f{}^2, ..., F_f{}^5\}$ and $F_m{}^i \in \{F_m{}^1, F_m{}^2, ..., F_m{}^5\}$, where the $F_f{}^i$ and $F_m{}^i$ are the output of the $i^{th}$ convolutional layer of $P_f$ and $P_m$. For feature propagation, the $F_f{}^i \in \{F_f{}^1, F_f{}^2, ..., F_f{}^5\}$ and $F_m{}^i \in \{F_m{}^L, F_m{}^{L+1}, ..., F_m{}^5\}$ are propagated to the SDCL decoder through skip connections. As exemplified in Fig. 1, when $L = 3$, the $F_m{}^1$ and $F_m{}^2$ are not propagated to the SDCL decoder.

## 2.2 Single-pass Deep Cumulative Learning (SDCL)

Cumulative learning is a cognitive process of cumulating knowledge for subsequent cognitive development [13]. By the name (single-pass deep cumulative learning), we aim to emphasize two characteristics of the SDCL decoder: (i) it can perform coarse-to-fine registration in a single pass (iteration) of the network, and (ii) the coarse-to-fine registration is performed as a cumulative learning process, where the knowledge (features and transformations) learned at former steps can be cumulated and facilitate later steps. This is different from the existing iterative methods as they usually learn separate features for each iteration, in which the knowledge learned at each step is collapsed into a warped image and is not cumulated for later steps.

Specifically, the SDCL decoder uses five successive $3 \times 3 \times 3$ convolutional layers to cumulate features. An upsampling layer is used after each convolutional layer, except for the last one, to increase the resolution of feature maps by a factor of 2. LeakyReLU activation with parameter 0.2 is used after each convolutional layer. These five convolutional layers can support up to 5-step coarse-to-fine registration. However, for $L$-step coarse-to-fine registration ($L < 5$), only the last $L$ convolutional layers are used for the $L$ registration steps, while the other $(5 - L)$ convolutional lay-ers are only used to cumulate features. As exemplified in Fig. 1, when $L = 3$, the first two convolutional layers are used to cumulate features, and the first registration step is performed after the third convolutional layer to produces the $\phi_1$. The $\phi_1$ is upsampled by a factor of 2 to be $\widehat{\phi_1}$, and the $\widehat{\phi_1}$ is used to warp the $I_m{}^2$. Then, the warped image $I_m{}^2 \circ \widehat{\phi_1}$ and the $\widehat{\phi_1}$ are fed into a convolutional layer to be leveraged at the second registration step. The second registration step also produces a transformation based on the cumulated features, but this transformation needs to be voxel-wisely added to the $\widehat{\phi_1}$ to derive the $\phi_2$. We repeat this process until the $\phi_L$ is derived. The $\phi_L$ is the final output of the NICE-Net, which can warp the $I_m$ to align with the $I_f$.

## 2.3 Unsupervised Training

The proposed NICE-Net is end-to-end trained using an unsupervised loss $\mathcal{L}$ without ground-truth labels. The loss $\mathcal{L}$ is defined as $\mathcal{L} = \sum_{i=1}^{L} \frac{1}{2^{(L-i)}} \left( \mathcal{L}_{sim}^i + \sigma \mathcal{L}_{reg}^i \right)$, where the $\mathcal{L}_{sim}^i$ is a similarity term that penalizes the differences between the warped image $I_m{}^i \circ \phi_i$ and the fixed image $I_f{}^i$, the $\mathcal{L}_{reg}^i$ is a regularization term that penalizes unrealistic transformations $\phi_i$, and the $\sigma$ is a regularization parameter.

As the local normalized cross-correlation (NCC) has been reported as a successful similarity metric in many deformable registration methods [8-12], we use negative NCC with window size $9^3$ as the $\mathcal{L}_{sim}^i$. The $\mathcal{L}_{reg}^i$ imposes L2 regularization on the $\phi_i$ to encourage its smoothness and also has a term $\mathcal{L}_{inv}^i$ to enhance its invertibility, which is defined as $\mathcal{L}_{reg}^i = \sum_{p \in \Omega} ||\nabla \phi_i(\boldsymbol{p})||^2 + \lambda \mathcal{L}_{inv}^i$ with the $\lambda$ as a regularization parameter. As the $\phi_i$ is not invertible at the voxel $p$ where the Jacobian determinant is negative $(|J\phi_i(p)| \leq 0)$ [14], we adopt the regularization loss proposed by Kuang et al. [15] as the $\mathcal{L}_{inv}^i$ to explicitly penalize the negative Jacobian determinants of $\phi_i$.

## 3 Experimental Setup

### 3.1 Datasets

We evaluated our NICE-Net with the task of inter-patient 3D brain MRI registration, and this task has been well benchmarked for registration with large deformations [8-11]. We used 2,760 T1–weighted brain MRI volumes for training, which were acquired from four public datasets: ADNI [16], ABIDE [17], ADHD [18], and IXI [19]. For validation and testing, we used two public datasets of brain MRI with manual segmentation: Mindboggle101 [20] and Buckner40 [21]. The Mindboggle101 dataset contains 101 MRI volumes. We randomly separate the dataset into 50 volumes for validation and 51 volumes for testing. The Buckner40 dataset consists of 40 MRI volumes and we used the dataset only for testing.

We performed brain extraction and intensity normalization for each MRI volume by FreeSurfer [21]. Then, each volume was affine-transformed and resampled to align with a MNI-152 brain template with 1mm isotropic voxels by FLIRT [22]. Finally, all volumes were cropped into 144×192×160 voxels.

### 3.2    Implementation Details

We implemented the NICE-Net using Keras with a Tensorflow backend on an Intel Core i5-9400 CPU and a 12 GB Titan V GPU. We used an ADAM optimizer with a learning rate of 0.0001 and a batch size of 1 to train the network for 100,000 iterations. At each iteration, two volumes were randomly picked from the training set as the fixed and moving images. A total of 200 image pairs, randomly picked from the validation set, were used to monitor the training process and to optimize hyperparameters. The $\sigma$ is set as 1 to ensure that the $\mathcal{L}_{sim}$ and $\sigma\mathcal{L}_{reg}$ have close values, while the $\lambda$ is set as $10^{-4}$ to ensure that the percentage of negative Jacobian determinants is no more than 0.05% (Table S1 in Supplementary Materials). We also trained a NICE-Net with $\lambda = 0$ to maximize the registration accuracy. Our code is available at `https://github.com/MungoMeng/Registration-NICE-Net`.

### 3.3    Comparison Methods

The proposed NICE-Net was compared to eight state-of-the-art registration methods, including two traditional methods, three non-iterative deep registration methods, and three iterative deep registration methods. The included traditional methods are SyN [23] and NiftyReg [24]. We ran them using cross-correlation as similarity measure with the parameters tuned on the validation set. The included non-iterative deep registration methods are Voxelmorph (VM) [4], Diffeomorphic Voxelmorph (DifVM) [5], and Dual-PRNet [11]. The included iterative deep registration methods are RCN [8], LapIRN [9], and ULAE-net [10]. For a fair comparison, we reimplemented all deep registration methods with Keras and used the same NCC loss as the similarity metric. We set $L = 3$ for the iterative deep registration methods (RCN, LapIRN, and ULAE-net), which makes them use out all the GPU memory (12 GB). Moreover, as the default VM and DifVM have fewer parameters than other methods, we increased their feature map channels to make them use out all the GPU memory as well.

### 3.4    Experimental Settings

We first compared the NICE-Net with the eight comparison methods for subject-to-subject registration. In this experiment, we set $L = 3$ for our NICE-Net, which is consistent with the RCN, LapIRN, and ULAE-net. Then, we performed an ablation study, where the NICE-Net with $\lambda = 0$ was evaluated with different $L \in \{1, 2, 3, 4, 5\}$. When $L = 1$, the NICE-Net only performs one-step registration, which means the SDCL decoder is not working and the SFL encoder has been degraded as a normal dual-path encoder without selective feature propagation.

A total of 200, 100 testing pairs were randomly picked from the Mindboggle101, Buckner40 testing sets for evaluation. The registration accuracy was evaluated by the Dice similarity coefficients (DSC) between fixed and warped segmentation masks. A two-sided $P < 0.05$ is considered to indicate a statistically significant difference. The transformation invertibility was evaluated by the percentage of negative Jacobian determinants (NJD). A lower NJD indicates a more invertible transformation.

## 4    Results and Discussion

The results of our NICE-Net and the comparison methods are shown in Table 1, and the qualitative comparison is shown in Fig. S1 in Supplementary Materials. Compared to the VM and DifVM, the iterative deep registration methods (RCN, LapIRN, and ULAE-net) achieved higher DSCs but had nearly double runtime as they performed iterative coarse-to-fine registration to handle large deformations. The Dual-PRNet also achieved higher DSCs than the VM and DifVM but cannot outperform the iterative deep registration methods. This suggests that the Dual-PRNet, although can realize non-iterative coarse-to-fine registration, is a non-optimal solution for large deformation registration. In the Dual-PRNet, the transformation produced at each registration step is based on the feature maps warped by its adjacent former step, while the knowledge learned at other former steps (except for the adjacent one) can hardly be leveraged. However, in our NICE-Net, coarse-to-fine registration is performed as a cumulative learning process, where the knowledge (features and transformations) learned at all coarser steps are cumulated and facilitate the registration at finer steps. Therefore, our NICE-Net ($\lambda = 0$) achieved significantly higher DSCs than all the comparison methods while only requiring similar runtime to the non-iterative methods. We noted that a recent study shows that the Dual-PRNet can be enhanced by computing local correlations between features or by joint learning with segmentation [25]. We anticipate that our NICE-Net can also benefit from these enhancements, and we will investigate its relative performance in our future study.

When $\lambda = 10^{-4}$, our NICE-Net achieved the lowest NJDs with a small degradation on DSCs. Compared to the NICE-Net ($\lambda = 10^{-4}$), the DifVM and LapIRN can achieve similar NJDs but had significantly lower DSCs; the ULAE-net can achieve similar DSCs but had much higher NJDs. These results demonstrate that our NICE-net can outperform the state-of-the-art iterative deep registration methods on all registration accuracy, registration speed, and transformation invertibility.

**Table 1.** Results of our NICE-Net and the comparison methods.

| Methods | Mindboggle101 | | Buckner40 | | Runtime (second) | |
|---|---|---|---|---|---|---|
| | DSC | NJD | DSC | NJD | CPU | GPU |
| SyN | $0.548^{*,\ddagger}$ | 0.26% | $0.577^{*,\ddagger}$ | 0.25% | 3793 | / |
| NiftyReg | $0.567^{*,\ddagger}$ | 0.34% | $0.610^{*,\ddagger}$ | 0.30% | 166 | / |
| VM | $0.558^{*,\ddagger}$ | 2.53% | $0.592^{*,\ddagger}$ | 2.22% | 3.85 | 0.395 |
| DifVM | $0.531^{*,\ddagger}$ | 0.04% | $0.574^{*,\ddagger}$ | 0.02% | 3.92 | 0.446 |
| Dual-PRNet | $0.586^{*,\ddagger}$ | 2.23% | $0.618^{*,\ddagger}$ | 2.13% | 4.47 | 0.467 |
| RCN | $0.592^{*,\ddagger}$ | 3.95% | $0.630^{*,\ddagger}$ | 4.02% | 6.75 | 0.692 |
| LapIRN | $0.596^{*,\ddagger}$ | 0.04% | $0.625^{*,\ddagger}$ | 0.03% | 6.52 | 0.624 |
| ULAE-net | $0.610^{\ddagger}$ | 2.00% | $0.640^{\ddagger}$ | 1.94% | 7.21 | 0.730 |
| NICE-Net ($\lambda = 10^{-4}$) | 0.608 | **0.03%** | 0.639 | **0.02%** | 4.17 | 0.423 |
| NICE-Net ($\lambda = 0$) | **0.621** | 2.01% | **0.649** | 1.96% | 4.17 | 0.427 |

**Bold**: the highest DSC and lowest NJD for each testing set are in bold.
*: $P < 0.05$, in comparison to the NICE-Net ($\lambda = 10^{-4}$).
$\ddagger$: $P < 0.05$, in comparison to the NICE-Net ($\lambda = 0$).

The results of the ablation study are shown in Table 2. The NICE-Net with $L = 1$ is regarded as the baseline, in which the SFL encoder and the SDCL decoder are not employed. The NICE-Net with $L > 1$ achieved higher DSCs than the baseline, which can be attributed to the use of our proposed SFL encoder and SDCL decoder. When the $L$ varied from 2 to 5, the DSC improved with a slight increase in runtime, which suggests that, in our NICE-Net, increasing registration steps results in higher registration accuracy while only adding a negligible computational load. This also means, if we set $L$ as 5, the NICE-Net can outperform the RCN, LapIRN, and ULAE-net by a larger margin while still keeping its advantage on registration speed. However, for the iterative deep registration methods, increasing $L$ means they have to cascade more networks or run the network for more iterations, which inevitably requires more GPU memory and further prolongs their runtime. We illustrate the results of the NICE-Net with $L = 5$ in Fig. 2. We found that the NICE-Net can perform finer registration after each step, gradually making the moving image $I_m$ closer to the fixed image $I_f$.

**Table 2.** Results of our NICE-Net with different $L$.

| NICE-Net with $L =$ | | 1 | 2 | 3 | 4 | 5 |
|---|---|---|---|---|---|---|
| Mindboggle101 | DSC | 0.565 | 0.601 | 0.621 | 0.624 | **0.626** |
| | NJD | 2.26% | 2.08% | 2.09% | 2.08% | 2.11% |
| Buckner40 | DSC | 0.599 | 0.629 | 0.649 | 0.652 | **0.654** |
| | NJD | 2.05% | 1.98% | 1.98% | 1.95% | 1.96% |

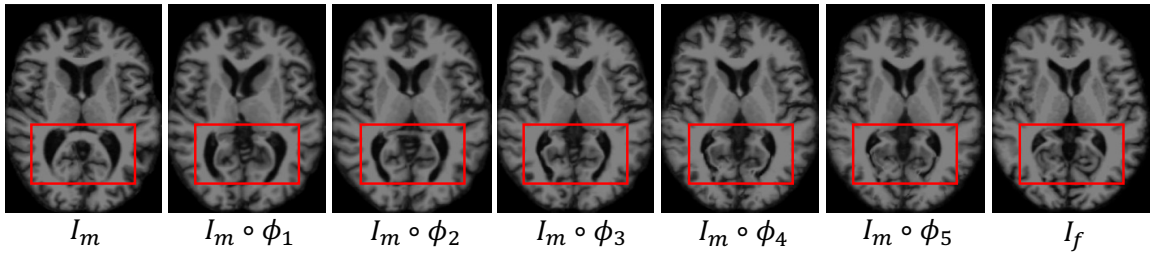

$I_m$     $I_m \circ \phi_1$     $I_m \circ \phi_2$     $I_m \circ \phi_3$     $I_m \circ \phi_4$     $I_m \circ \phi_5$     $I_f$

**Fig. 2.** Registration results of the NICE-Net with $L = 5$. From life to right are the moving image, the images warped by 5 registration steps, and the fixed image.

## 5 Conclusion

We have outlined a Non-Iterative Coarse-to-finE registration Network (NICE-Net) for deformable image registration. Unlike the existing iterative deep registration methods, our NICE-Net can perform coarse-to-fine registration with a single network in a single iteration. The experimental results show that the proposed NICE-Net can outperform the state-of-the-art iterative deep registration methods on both registration accuracy and transformation invertibility while only requiring similar runtime to non-iterative registration methods.

**Acknowledgement.** This work was supported in part by Australian Research Council (ARC) grants (IC170100022 and DP200103748).